\documentclass[10pt,twocolumn,letterpaper]{article}

\usepackage{cvpr}              

\usepackage{graphicx}
\usepackage{amsmath}
\usepackage{amssymb}
\usepackage{booktabs}
\usepackage{flushend}

\usepackage{subcaption}
\usepackage{booktabs}
\usepackage{multirow}
\usepackage{mwe}
\makeatletter
\@namedef{ver@everyshi.sty}{}
\makeatother
\usepackage{tikz}
\usepackage{comment}
\usepackage{amsmath,amssymb} 
\usepackage{color}

\newcommand*\mycirc[1]{%
   \begin{tikzpicture}[baseline=(C.base)]
     \node[draw,circle,inner sep=0.5pt, ](C) {\tiny #1};
   \end{tikzpicture}}

\usepackage[pagebackref,breaklinks,colorlinks]{hyperref}

\usepackage[capitalize]{cleveref}
\crefname{section}{Sec.}{Secs.}
\Crefname{section}{Section}{Sections}
\Crefname{table}{Table}{Tables}
\crefname{table}{Tab.}{Tabs.}

\def\cvprPaperID{9978} 
\def\confName{CVPR}
\def\confYear{2023}

\begin{document}

\title{Detection of out-of-distribution samples using binary neuron activation patterns}

\author{Bartłomiej Olber$^{1, 2}$, Krystian Radlak$^{1,2}$, Adam Popowicz$^{2}$, \\ Michal Szczepankiewicz$^{3}$, Krystian Chachuła$^{1}$\\
$^{1}$Warsaw University of Technology $^{2}$Silesian University of Technology $^{3}$NVIDIA \\
{\tt\small \{bartlomiej.olber.stud, krystian.radlak, krystian.chachula\}@pw.edu.pl }\\ 
 {\tt\small adam.popowicz@polsl.pl  msz@nvidia.com
}
}

\maketitle

\begin{abstract}
   Deep neural networks (DNN) have outstanding performance in various applications. Despite numerous efforts of the research community, out-of-distribution (OOD) samples remain a significant limitation of DNN classifiers. The ability to identify previously unseen inputs as novel is crucial in safety-critical applications such as self-driving cars, unmanned aerial vehicles, and robots. 
   Existing approaches to detect OOD samples treat a DNN as a black box and evaluate the confidence score of the output predictions. Unfortunately, this method frequently fails, because DNNs are not trained to reduce their confidence for OOD inputs.
In this work, we introduce a novel method for OOD detection.  Our method is motivated by theoretical analysis of neuron activation patterns (NAP) in ReLU-based architectures. The proposed method does not introduce a high computational overhead due to the binary representation of the activation patterns extracted from convolutional layers.  The extensive empirical evaluation proves its high performance on various DNN architectures and seven image datasets. 

\end{abstract}

\section{Introduction}
\label{sec:intro}
Even the most efficient deep neural network (DNN) architectures, designed for image recognition tasks, cannot ensure that they will not malfunction during their operation. Thus, deployment of those safety-critical applications, such as in self-driving cars, unmanned aerial vehicles, and robots, is still an unresolved problem \cite{FalciniLami2017,KhanAmin2021}. The use of safety mechanisms, such as runtime monitors, is a viable strategy to keep the system in a safe state despite of DNN failure. The design and development of such monitors in the context of safety-critical applications is a significant challenge \cite{ForsbergLinden2020,TerrosiStrigini2022}. Therefore, it is required to define robust metrics that can allow to detect and control of DNN's failures at runtime and mitigate potential hazards caused by their performance limitations. 

\par DNNs are trained over a set of inputs sampled from real-world scenarios. However, due to the large variation of the input images, the training dataset cannot contain all possible variants of input samples. Although it is expected that the trained models can perform well on unknown inputs, especially those that are similar to the training data, it cannot be guaranteed that they will perform well for OOD noisy samples that present the objects not considered before \cite{DBLP:journals/corr/abs-2102-12967}. While DNN training techniques should allow a network to achieve high generalization capabilities, it is also crucial to ensure the dependability of safety-critical systems to train a model so that any outlying input will result in low confidence of the network's decision. 

\par The fundamental challenge to ensure the safety of DNNs is to estimate if a given input sample comes from the same data distribution for which a DNN was trained. This is very hard to estimate because the network usually extrapolates its decision while receiving new image samples. Another cause of the incorrect recognition of outlying samples can be the distributional shift of input data over time (e.g. the time-dependent variations of an object's appearance) \cite{HellHinz2021}.


\par In the vast literature, this problem has been formulated as a problem of detecting whether input data are from an in-distribution (ID) or out-of-distribution (OOD). This has been studied for many years and discussed in the following aspects: sample rejection, anomaly detection, open-set samples recognition, familiar vs unfamiliar samples or uncertainty estimation \cite{BendaleBoult2016, KendalGal2017,PangShen2021 }. 


\par In this work, we present a novel algorithm for the identification of OOD image samples. In our method, we extract the binary neuron activation patterns on various hidden layers of a DNN and compare them with the ones collected in the training procedure. By measuring the Hamming distances between  extracted binary patterns of any test sample and the patterns extracted during the training,  we can identify OOD samples. 

The main contributions of this paper are the following:
\begin{itemize}
    \item We introduce NAP - an algorithm that extracts binary patterns from both fully connected and convolutional layers and estimates a classifier's predictive uncertainty based on the patterns. The proposed method outperforms state-of-the-art OOD detection methods. Moreover, the algorithm is straightforward, making it simple to incorporate into existing DNN architectures.
    \item We provide an extended empirical evaluation comparing the impact of the activation patterns collected from different layers of DNN which may inspire future research in this area.
    \item We publish the largest evaluation framework for OOD detection. This framework contains 17 OOD methods (including the proposed NAP-based method) that can be directly tested on two state-of-the-art DNN architectures and 7 datasets allowing for simple extension of the framework for new methods, architectures, and datasets.\footnote{\href{https://github.com/safednn-group/nap-ood}{https://github.com/safednn-group/nap-ood}}
\end{itemize}



\section{Related work}
In the literature, a plethora of algorithms designed for the detection of OOD samples for DNNs have been proposed \cite{yang2021generalized}. 
One of the most promising groups of methods is based on a hidden layers' activation analysis. This type of method has found its use in specialized problem domains such as cybersecurity malware classification \cite{DBLP:journals/corr/abs-1903-04717}, autonomous driving safety monitors \cite{Henzinger2020OutsideTB,cheng2018runtime, 10.1007/978-3-030-88494-9_14} or in explaining patient medical diagnosis decision \cite{stano2020explaining}. Also, from a domain-agnostic point of view, there are several topics into which the activation analysis can be incorporated. 

\par In this group, some methods primarily derive OOD scores based on an analysis of activations. Authors in \cite{stano2020explaining} focus on explainable AI. They derive binary patterns from the activation layers of a DNN, then compute the Hamming distance to find the most perceptually similar
and dissimilar known ground-truth images which are then used to explain the decision made by the DNN model. The authors of \cite{DBLP:journals/corr/abs-1911-12780} expressed a need for detecting rare subclasses in datasets because they might reduce a DNN's performance. They proposed a method based on a simple statistical analysis of penultimate layer activations dealing with the issue. In the works \cite{Henzinger2020OutsideTB}, \cite{10.1007/978-3-030-88494-9_14}, \cite{cheng2018runtime}, the activation outputs of the hidden layers are processed to build a safety monitor. The paper \cite{Henzinger2020OutsideTB} shows the utilization of the k-means clustering and abstraction boxes. In \cite{10.1007/978-3-030-88494-9_14} the authors build a GMM for every neuron independently and establish safety intervals for every neuron based on the mean and the standard deviation of the previously built models. Moreover, they use a voting system to improve the decision's correctness. The method given in \cite{cheng2018runtime} makes use of the binary activation patterns and the Hamming distance for OOD sample detection. However, the presented approach has significant efficiency issues that make the approach unscalable to any modern DNN architecture. In \cite{meronen2020stationary}, the authors proved that activation function choice itself is of great importance for OOD uncertainty calibration methods. They propose the Matérn activation function which improves the Monte Carlo Dropout \cite{gal2016dropout} uncertainty estimation. 

\par There are also uncertainty estimation methods that operate strictly on internal activations. The work \cite{meegahapola2019prior} brought up an interesting technique of uncertainty measurement that leverages aggregated, statistical properties of hidden-layers activations that capture cross-class discrimination capability. The approach given in \cite{postels2021hidden} is based on the density of latent representations. Applying reconstruction loss to the density yields a high-quality uncertainty estimation that enables OOD detection. The authors, like in the papers \cite{postels2019samplingfree} and \cite{ovadia2019trust}, conclude that there are benefits from using the output of various hidden layers (shallow, medium, deep, etc.) instead of focusing only on the last layer of a DNN.


\par In the paper describing the ODIN method \cite{liang2020enhancing}, authors proved both empirically and theoretically that temperature rescaling \cite{guo2017calibration} and input perturbation \cite{goodfellow2015explaining} can provide well-calibrated softmax scores on which the OOD detector can be constituted. In the work \cite{lee2018simple}, authors measure the probability density of a test sample on feature spaces and take Mahalanobis distance with respect to the closest class-conditional distribution. The method achieves outstanding OOD detection efficiency. The previous work of \cite{hendrycks2019deep} shows the Outlier Exposure technique which tackles a widely encountered problem of classifier overconfidence on OOD samples. They propose to incorporate an OOD penalty into a loss function during the training phase of a DNN.
Also, the algorithms based on Helmholtz energy score \cite{928a56b7d6f1473e930f282a0c4b534e} have recently become very popular, with the EnergyOOD \cite{liu2021energybased} and its upgraded versions - ReAct \cite{sun2021react} and ASH \cite{extremely} establishing the state-of-the-art performance level.  

A completely different approach is introduced in the work of \cite{huang2021importance} where the authors focus on a gradient space as adequate for the OOD detection task. State-of-the-art anomaly detection methods \cite{reiss2021meanshifted}, \cite{reiss2021panda} adapt pretrained features with a center loss and combine it with contrastive learning. Although the focus was put on one-class classification, the introduced methods are effective OOD detectors considering a known distribution as the normal class and a semantic-shifted distribution as the OOD samples.

\section{Binary Neuron Activation Patterns}

In comparison to simple linear classifiers, DNNs have a much higher expressive power arising from non-linear activation functions between linear layers. One of the most significant is the ReLU non-linearity introduced in \cite{NairHinton2010}. ReLU-based neural networks have led to state-of-the-art results in many fields, but the computational structure of the ReLU activation is straightforward. ReLU non-linearity has only two states: "inactive" and "activated". The expressivity of a fully trained network depends on the ability to select appropriate activations for different inputs. It was theoretically proven in \cite{HaninRolnick2019} that a DNN uses much fewer activation patterns during training than theoretically possible. Moreover, in \cite{Hartmann2021}, the authors presented that the non-linear structure converges bottom-up (lower layers stabilize first) during training in ReLU-based architectures and shows stable distribution over pattern set changes. 

\par These findings naturally inspire a simple, yet surprisingly effective, method in which the comparison of NAPs allows prediction uncertainty to be measured in the inference stage. In this approach, each layer $l$ in a DNN generates an activation vector $A_l$ based on the processed input data. We define the
activation pattern $A_l$ on layer $l$ of DNN with a vector of trainable parameters $\theta$ for an input $x_i,\; i=1,\ldots,n$ as  an assignment of a sign to each neuron $z_j, \; j=1,\ldots,m$
\begin{gather}
    A_l(x_i, \theta) := \{ a_{z_j}, \; j=1,\ldots,m\} \in \{0,1\}, \textrm{where} \\
     a_{z_j}=\begin{cases}
    1, & \text{if $z_j(x_i,\theta)>=0$}.\\
    0, & \text{otherwise}.
  \end{cases}
\end{gather}

\begin{figure*}
    \centering
    \includegraphics[width=0.88\textwidth]{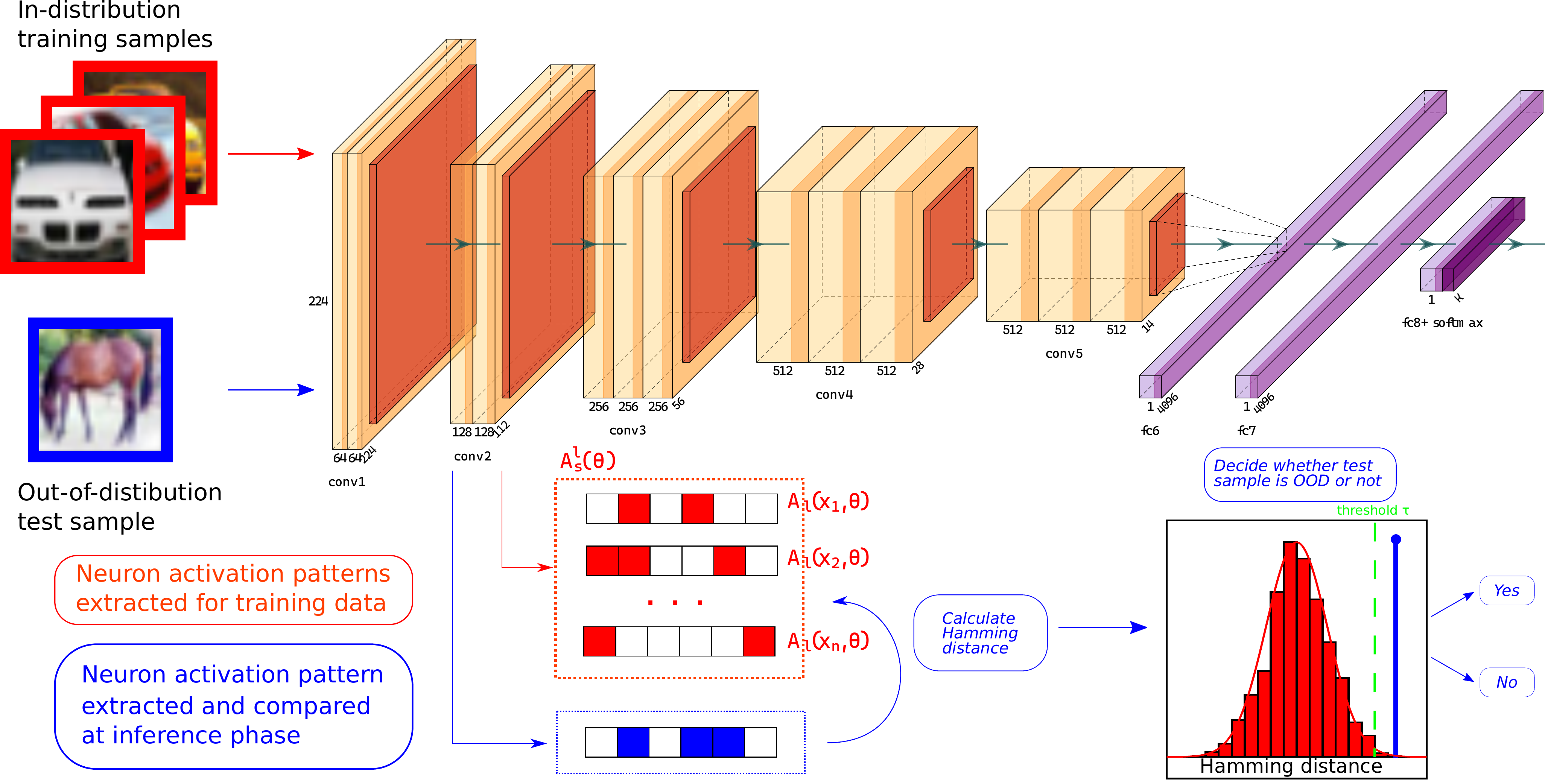}
    \caption{The idea of the proposed OOD detection method using binary neuron activation patterns.}
    \label{fig:nap}
\end{figure*}

Each unit of the activation vector can be used as a feature to represent the part of the DNN that was activated by a certain sample from the training dataset. We collect those binary patterns $A_l(x_i, \theta)$ for all samples during the training procedure and we compare the binary pattern extracted from a test sample in order to decide whether a tested sample is OOD or not. Fig \ref{fig:nap} shows a simplified diagram summarizing our approach. In the presented chart, the red color indicates NAPs extraction for the training dataset, while the blue refers to the testing process of whether a new sample comes from the known distribution.

While extraction of NAP for a feedforward network and a fully connected layer is straightforward, the method of NAP extraction for a convolutional layer is presented in section 3.2.


\par For ReLU-based architectures, it can be noticed that for each neuron $z_j$ of every layer $l$, the gradients applied to its weights are null if the neuron is not activated.\footnote{The proof is provided in the supplementary material} This property proves that simplified binary patterns may be a sufficient representation of ID samples and we do not need to operate on network activation values.


\subsection{Detecting OOD samples using NAP}
\par The main expectation is that a trained model should be more certain when predicting the output for a sample similar to the ones seen during the training phase. Therefore, the similarity measured by the distance between the test sample and training samples should be directly connected with the model's uncertainty. In the proposed approach, we compute the smallest Hamming distance between the NAP of a given test sample and samples extracted from the training set. This distance for test sample $v$ can be defined as follows:
\begin{equation}
d_{min}(v,\theta) = \min\{d(v, u),\forall u \in A_{s}^l(\theta)\},
\end{equation} 
where $ d(v, u) $ denotes the Hamming distance between binary vectors $v$ and $u$ (the Hamming distance between two binary patterns is equal to the number of different bits between them); $A_{s}^l(\theta) =\{ A_l(x_i, \theta), \; i=1,\ldots, n\}$ is the set of all NAP extracted for the training database for a selected layer $l$.

Computing the Hamming distance for a test sample requires creating a set of binary activation patterns for all known training images. Depending on the factors, such as the number and dimensionality of activation patterns, there are two efficient ways to compute the smallest distance. In the first method, for every incoming pattern, the XOR operation is performed with all collected patterns in the training stage. The example for which the XOR result has the lowest number of positives is selected as the most similar sample.
    Although this method requires computing and comparing the results for every pattern from the training set, modern libraries can parallelize simple XOR operations very well. Still, the number of operations is rather large because a typical dataset contains tens or even hundreds of thousands of samples, but it still can be efficiently calculated on a GPU.
     \par In the second method, patterns from the training set need to be stored in a data structure that is dedicated to quickly solving the nearest neighbor search problem (e.g., the ball tree or binary hashing-based approximate nearest neighbor search methods \cite{CaoQi2018}). These methods provide fast query speed and drastically reduced storage, which allows us to efficiently apply the proposed NAP method even for very large training datasets. In this work, we used the first approach, which provides sufficient efficiency to implement a NAP-based detector as a runtime monitor in real-time applications, which was proven in the Experiments section.

\subsection{NAP extraction from convolutional layers}
Convolutional layers are the backbone of most modern DNN architectures utilized in computer vision tasks. They forward multi-dimensional feature representations from one layer to another until the features are eventually represented by a one-dimensional vector. Extracting relevant activation patterns out of a convolutional layer is more complex than for fully connected layers. One of the main novelties in our approach is that we are proposing how to extract the binary activation patterns for convolutional layers in DNN. Simple flattening of multi-dimensional layers would lead to NAPs of an overwhelming length. 

According to the best of our knowledge, it has not yet been proposed how to extract binary activation patterns for convolution layers. We realize that the presented method is only one of many possible ways to generate such patterns and deeper investigation on this topic will be the subject of future research work.

To combine the potential of hidden convolutional layers but also to mitigate the problem of the high dimensionality of data, we propose to perform adaptive pooling of all channels in a convolutional layer into a single value. We compress outputs from all channels in an extracted layer into a vector by such a procedure. The proposed pooling with a $t$-type aggregator  of a convolutional layer $X$ ($H\times W$ elements, $C$ channels) into a $C$-elements vector $Y$ can be expressed by the following formula:
\begin{equation}
    pool_{t}(X) \rightarrow Y.
\end{equation}
In this research, we evaluated two types of pooling: average pooling and max pooling. In the performed experiments, we proved that max pooling provides better results for OOD detection, and therefore, we recommend it as the default method for extraction of binary activation patterns from convolutional layers. The intuition behind this type of pooling is that we are extracting the strongest pattern from each channel, which allows us to remove most of the non-important values from each layer.

Next, the values received after pooling need to be converted into a binary NAP. For this purpose, we propagate the vector through a configurable, adaptive activation function that binarizes the values using individual thresholds for each layer. As a threshold, we use the $p$-percentile of values for a given activation pattern. for each layer. As a threshold we use the p-percentile of values in a given activation pattern. This results in zeroing the elements of the vector that represents the least significantly activated channels. Practically, this allows us to automatically choose the most important subset of channels for every layer based on the activation magnitude. 

\begin{figure}
    \centering
    \includegraphics[width=0.42\textwidth]{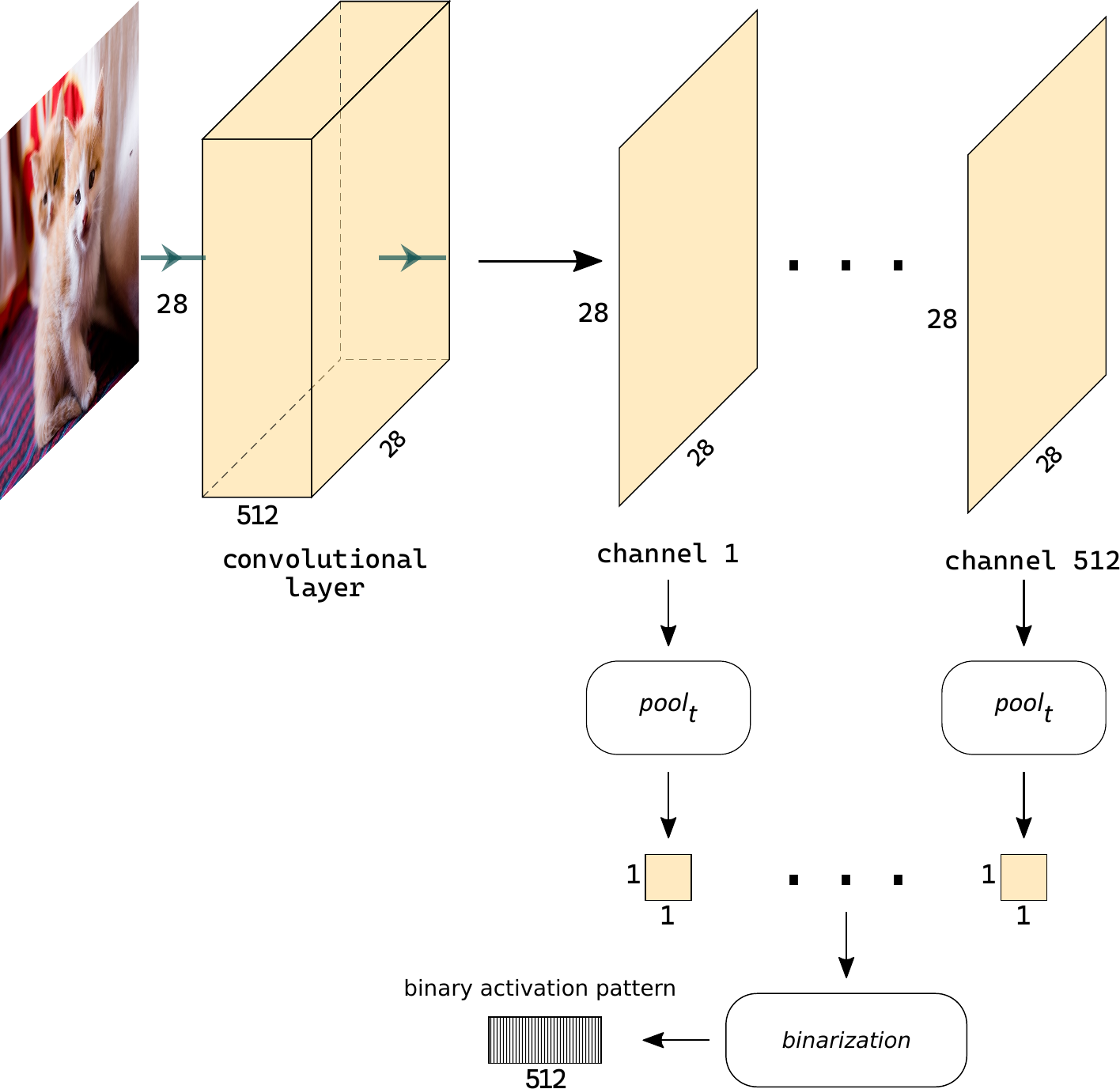}
    \caption{Proposed approach for extraction of a binary activation pattern from a convolutional layer.}
    \label{fig:extracting.png}
\end{figure}

An example of extraction of a binary activation pattern for the convolutional layer is depicted in Fig. \ref{fig:extracting.png}. It can be seen that the convolutional layers are first pooled for every channel in a selected convolutional layer and then binarized. The general formula for obtaining an activation pattern for a network consisting of a total number of $N$ channels in the convolutional layers and $M$ elements in the neural layers region of a network can be expressed as follows:
\begin{equation*} 
\begin{aligned}
A_l(x_i, p) = [\alpha\{pool_{t}(V_{0}),P_{p}^{0}\}, ...,\alpha\{pool_{t}(V_{N}),P_{p}^{N}\}, \\
\alpha\{V_{N+1},\  P_{p}^{N+1}\},...,\alpha\{V_{N+M},\  P_{p}^{N+M}\}],
\end{aligned}
\end{equation*} 
\begin{equation}
\alpha \{x, p\} = \begin{cases}
  1 \ \  x > p \\ 
  0 \ \ x <= p ,
\end{cases}
\end{equation}
where $\alpha$ is the binarization function, $V_i$ is the $i$-th value in the activation vector (before binarization), $P_p^N$ is the value of $p$-percentile (within an assumed dataset) at $N$-th position in the activation vector. 


\subsection{Uncertainty threshold estimation}
In order to decide whether a new sample is an OOD or not, it is required to define a decision criterion.  
In the proposed NAP approach,  we use the Hamming distance between a binary pattern extracted for a test sample and all binary patterns extracted from samples collected in the training dataset. If the Hamming distance to the nearest binary pattern is higher than the assumed threshold $\tau$, then a sample is classified as an OOD. 
\par Let the corresponding sets of distances be called $D_{in}$ and $D_{out}$, for ID and OOD samples, respectively. The distance $d_i$ for $i$-th image sample is calculated to the nearest neighbor in an ID set as follows:
\begin{equation}
d_{i} = \left \{ \min_j(HD(A(x_i,\theta),A(x_j,\theta)))  : j \in A_{in},i\neq j \right \} ,
\end{equation}
where $i$ and $j$ are sample numbers, $A_l(x_i, \theta)$ is the activation pattern of $i$-th sample, $A^l_s(\theta)$ is the set of all samples in an ID set, and $HD(a,b)$ is the Hamming distance between samples $a$ and $b$.
\par The separation threshold was estimated using an exhaustive search approach in which the decision threshold $\tau$ is established as a value that minimizes the intra-class variance between the Hamming distances distributions calculated between NAP for training and a validation OOD dataset.

\subsection{Method auto-configuration}

The proposed approach needs to be tuned depending on the following parameters: (1) layers of a neural network, (2) $p$-percentile values, or (3) type of pooling $t$. To distinguish most precisely the ID pattern from the OOD pattern for a wide range of DNN models and various dataset distributions. Finding an optimal combination of these three parameters is a model-wise or even a layer-wise problem. Our strategy is to divide the auto-configuration problem into the following two sub-problems: (1) layer-wise optimization of parameters $p$ and $t$, (2) model-wise specification of most relevant layers. The optimization process comes down to selecting the described parameters based on the OOD detection results obtained using the validation set $D_v$.

\par First, we grid-search the best $p$ and $t$ for every activation layer in a model. Each layer is analyzed independently for its OOD detection efficiency in this step. In our initial experiments, we noticed that there is no universally optimal layer to be distinguished.  Depending on network architecture and training dataset some layers perform better than others. Therefore, we extract NAPs from the $k$ layers that yield the highest OOD detection accuracy on validation datasets, where $k$ is an arbitrary odd number. Then, we create a majority voting classifier based on the Hamming distances calculated for the $k$ layers. In our experiments, $k$ is another parameter that needs to be tuned for the proposed NAP-based OOD detector.  

\noindent \textbf{Optimization criteria} \label{criteria}
In order to find optimal hyperparameters of the NAP method, it is required to define the proper optimization criteria. In this work, we analyzed two of them: (1) OOD validation accuracy and (2) distance threshold $\tau$ (lower is better).  Whereas the justification for the accuracy criterion is obvious, a justification for the threshold criterion might not be. We follow the intuition that a low threshold is more robust to a possible shift between validation and test distributions than a high threshold. The third hybrid criterion is the combination of them both, which tries to get a high validation accuracy while maintaining a reasonably low $\tau$. In our experiments, we apply these criteria in the following scheme: 
\begin{enumerate}
    \item For each layer of a network, choose the best $p$ parameter by the hybrid criterion (in this step, the criterion is constant).
    \item Choose the optimal pooling type $t$ for each layer and its $p$ parameter chosen in the previous step (here, we experiment with different criteria).
    \item Choose $k$ layers that yield the best validation accuracy for parameters chosen in previous steps (criterion is constant).  
\end{enumerate}
Attained $p$, $t$, $\tau$, and $k$ parameters form together a NAP configuration. Note that the $p$ and $\tau$ parameters are tuned for each layer, but $t$ is tuned only for convolutional layers. 



\subsection{Combining distances of multiple layers} 
\label{combining}
Assuming that the algorithm has established the configuration and we can obtain the Hamming distances for binary patterns from numerous layers, these patterns can be combined into an ensemble decision-making algorithm that will classify a test sample as an OOD. In this work, we compare two decision criteria. The first decision algorithm (vote scheme 1) combines the Hamming distances into a single uncertainty score according to the following definition: 
\begin{equation}
        \label{eq:dist1}
        score = \sum_{l \in L} d_{l}^{scaled} - \tau_{l}^{scaled} \\
\end{equation}
        where $d_{l}^{scaled}$ is a scaled nearest Hamming distance obtained from the $l$-th layer; $\tau_{l}^{scaled}$ is a scaled Hamming distance threshold established for the $l$-th layer; $L$ is a set of chosen layers to monitor.
    
       

The second decision algorithm is a simple majority voting classifier (vote scheme 2), in which each layer votes independently whether a sample is OOD or not. The final decision is a mode of $k$ votes. Which imposes the $k$ parameter to be an odd number.

\section{Experiments}
\noindent
\textbf{OOD evaluation protocol}
We evaluated the proposed binary NAP for the OOD detection task following the methodology proposed in the OD-test framework \cite{ShafaeiSchmidt2019}. The assumption that an OOD detector challenged with real-world tasks would be exposed to samples belonging only to the two distributions (ID and OOD) can be tricky. An OOD detection method might not be robust enough to be engaged in a safety-critical environment unless one can provide a dataset that covers all possible semantic spaces. Shafaei et al. \cite{ShafaeiSchmidt2019} introduced an OOD test, which is a three- (instead of two-) dataset
evaluation methodology for the purpose of assessing OOD methods more reliably. It was proven that the results obtained with the three-set evaluation protocol have higher confidence and are more reliable.

\begin{figure}
    \centering
    \includegraphics[height=3.4cm]{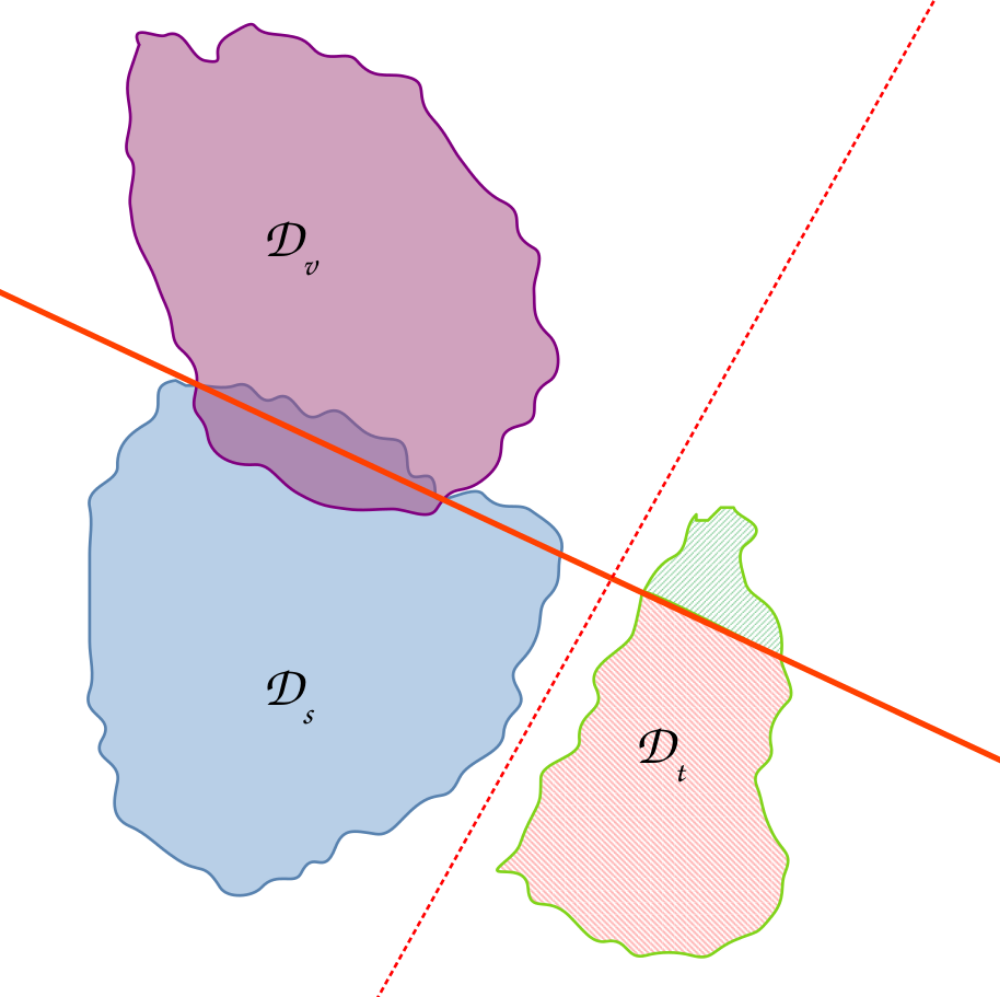}
    \caption{Fallacy of the two datasets OOD evaluation. The established threshold between source distribution $D_{s}$ and validation distribution $D_{v}$ could perform poorly if unknown test distribution $D_{t}$ differs from $D_{v}$. }
    \label{fig:od-test}
\end{figure}

In this evaluation procedure, the data are split into three sets: source training $D_{s}$, validation $D_{v}$, and test $D_{t}$ subsets. The original, ID images are included only in $D_{s}$, whereas the OOD samples are divided into two distinctively different groups: $D_{v}$ and $D_{t}$. The datasets are presented in Fig. \ref{fig:od-test}. The reliable OOD detector should be trained to discriminate $D_{s}$ from $D_{v}$, but its generalization capability to identify OOD should be performed on the $D_{t}$ against $D_{s}$  differentiation task.

\par
\noindent
\textbf{Models and datasets} In our evaluation, we used pre-trained VGG-16 \cite{simonyan2015deep} and ResNET-50 \cite{he2015deep} and seven datasets (MNIST \cite{LeCun2005TheMD}, FashionMNIST \cite{xiao2017fashionmnist}, NotMNIST, CIFAR10, CIFAR100 \cite{Krizhevsky2009LearningML}, STL10 \cite{article} and TinyImagenet) for the evaluation.  Each method under the OOD evaluation protocol is tested over the same set of
308 experiments consisting of all compatible triplets of datasets. As $D_s$, $D_v$, and $D_t$, we use all possible combinations from the above set of seven datasets. \par \noindent
\textbf{Metrics} Aside from accuracy (only this metric was calculated in the original OD-test framework), we used the area under the receiver
operating characteristic curve (AUROC) as a main evaluation metric since it is the appropriate metric for two-class discrimination tasks with balanced data. \newline
\noindent \textbf{Methods} We compared our methods with several state-of-the-art OOD detection methods to create the largest OOD evaluation framework. In our comparison, we used the methods implemented in the OD-test framework \cite{ShafaeiSchmidt2019}: {\it OpenMax} \cite{bendale2015open},  {\it ScoreSVM} \cite{Cortes2004SupportVectorN}, {\it LogisticSVM}, {\it BinClass}, {\it PbThreshold} \cite{hendrycks2018baseline}, {\it ODIN} \cite{liang2020enhancing}, {\it ReconstThreshold}, {\it K-NNSVM}, {\it K-MNNSVM}, {\it K-BNNSVM}, {\it K-VNNSVM}
\cite{kingma2014autoencoding} (where K denotes the number of nearest neighbors), {\it MCDropout} \cite{gal2016dropout} and {\it DeepEnsemble} \cite{lakshminarayanan2017simple}. Additionally, we integrated recently proposed state-of-the-art OOD detection methods: {\it Mahalanobis} \cite{lee2018simple}, {\it OutlierExposure} \cite{hendrycks2019deep}, {\it Energy} \cite{liu2021energybased}, {\it ReAct} \cite{sun2021react}, {\it GradNorm} \cite{huang2021importance}, {\it MeanShiftedAD} \cite{reiss2021meanshifted} and {\it ASH-B} \cite{extremely}. For all methods, we used the parameter configuration provided in official papers' implementations. 



\noindent \textbf{Adaptive configuration strategies}
In the proposed method, we evaluated several auto-configuration strategies and ways to score distances out of multiple layers. In Tab. \ref{table:Od-test results}, we present the final results for hyperparameters optimization for the analyzed network architectures using the accuracy metric. The hyperparameters included in the ablation study are as follows: the choice of the optimization criterion, number of voting layers, and OOD decision-making algorithm for the multiple voting layers. In the superscript brackets, we annotated the best configurations for each network architecture as NAP$_1$ and NAP$_2$, respectively. 
These configurations are used in comparison with state-of-the-art OOD detectors in the next subsection. Additionally, more results documenting the impact of NAP hyperparameters on OOD detection performance are presented in the supplementary materials.

\setlength{\tabcolsep}{4pt}
\begin{table}[!htb]
\caption{OD-test performance for selected hyperparameters. The \textbf{bold result} denotes the setups included in the OD-test comparison. 
}
\label{table:Od-test results}
\scriptsize
\begin{center}
\begin{tabular}{llcll} 
\hline \hline\noalign{\smallskip}
Model& Criterion & \#k - nb of layers  & $\textrm{Vote scheme } {1}$ & $ \textrm{Vote scheme } {2}$ \\
\noalign{\smallskip}
\hline \hline
\noalign{\smallskip}
& & & \multicolumn{2}{c}{Accuracy}  
\\ \noalign{\smallskip} \cline{4-5} \noalign{\smallskip}
\multirow{8}{*}{VGG} &\multirow{4}{*}{valid. acc.} & 3  & 81.39\% & 79.57\% \\
& & 5  & 81.94\% & 80.92\% \\
& & 7  & 82.01\% & 82.35\% \\
& & 9  & \textbf{82.23\%}$^{[NAP_1]}$ & \textbf{83.26\%}$^{[NAP_2]}$ \\
\cmidrule{2-5}
& \multirow{4}{*}{threshold} & 3  & 79.74\% & 80.21\%  \\
& & 5  & 79.55\% & 81.09\%  \\
& & 7  & 79.87\% & 81.72\%  \\
& & 9  & 79.91\% & 82.18\%  \\
\midrule
\multirow{8}{*}{ResNET} &\multirow{4}{*}{valid. acc.} & 3  & 79.05\% & 77.75\%  \\
& & 5  & 79.03\% & 77.46\%  \\
& & 7  & 79.15\% & 77.11\%  \\
& & 9  & \textbf{79.15\%}$^{[NAP_1]}$ & 77.50\%  \\
\cmidrule{2-5}
& \multirow{4}{*}{threshold} & 3  & 79.38\% & 79.54\%  \\
& & 5 & \textbf{79.82\%}$^{[NAP_2]}$ & 78.41\%  \\
& & 7  & 78.71\% & 76.72\%  \\
& & 9  & 77.95\% & 76.42\%  \\

\hline
\end{tabular}
\end{center}
\end{table}
\setlength{\tabcolsep}{1.4pt}

\noindent \textbf{Time efficiency} All the experiments were performed in Pytorch v1.8 environment and Python v3.9 client, running on a PC with Intel\textsuperscript{\mycirc{R}} Core\textsuperscript{TM} i7-3930K CPU 3.20 GHz and GeForce GTX 1080 Ti. All methods were integrated into the same framework and computation times were compared on the same setup. 
 A summary of time efficiency for evaluated methods is presented in Tab. \ref{table:time efficiency}. The execution time of the NAP method is comparable to other state-of-the-art methods such as ODIN (10 and 20 ms for VGG and ResNET architectures) and significantly faster than the most efficient Mahalanobis method, allowing it to be used in real-time applications.
 
\begin{figure*}[!htb]
\centering
   \begin{subfigure}{0.48\textwidth}
        \centering
    \includegraphics[width=\textwidth]{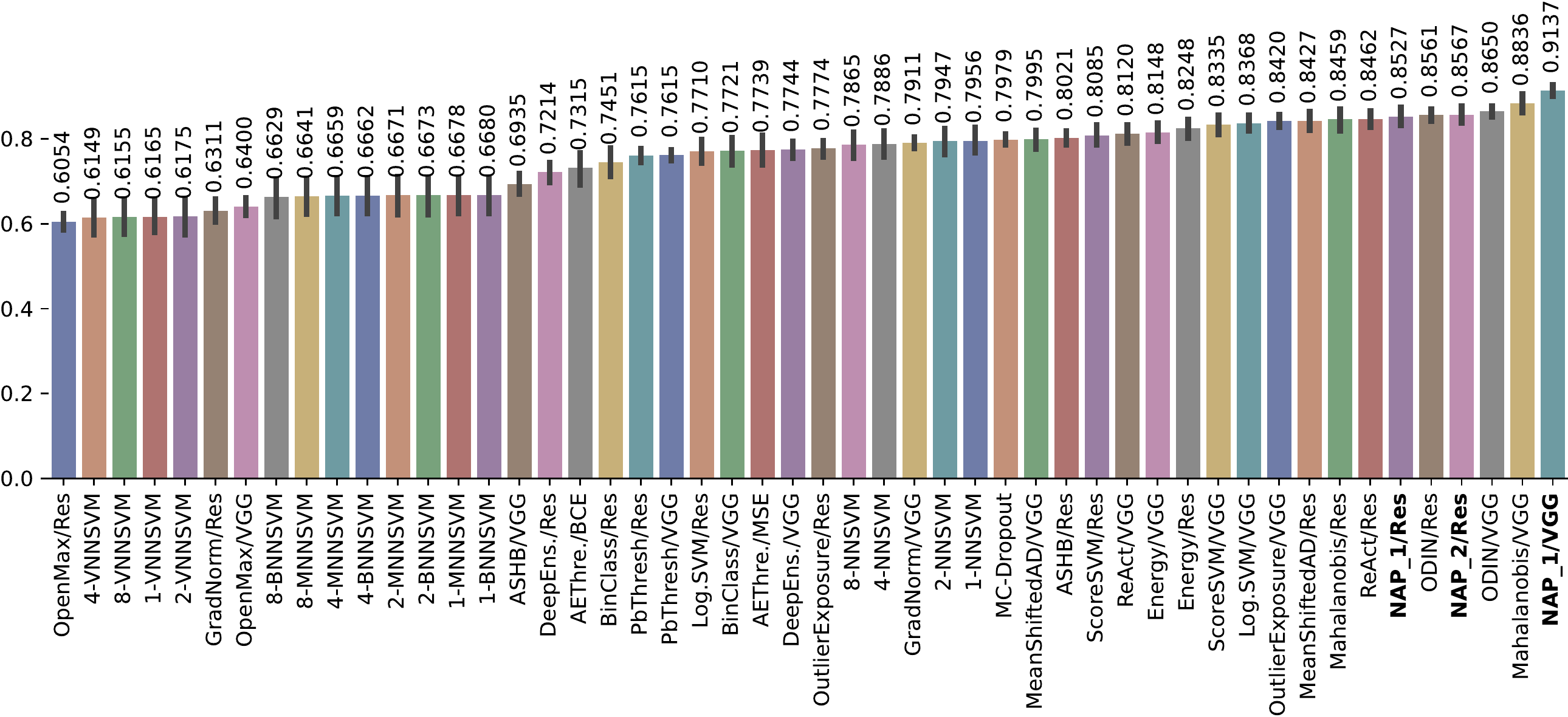}
    \caption{AUROC}
    \label{fig:odtest auroc}
\end{subfigure}
\begin{subfigure}{0.48\textwidth}
        \centering
    \includegraphics[width=\textwidth]{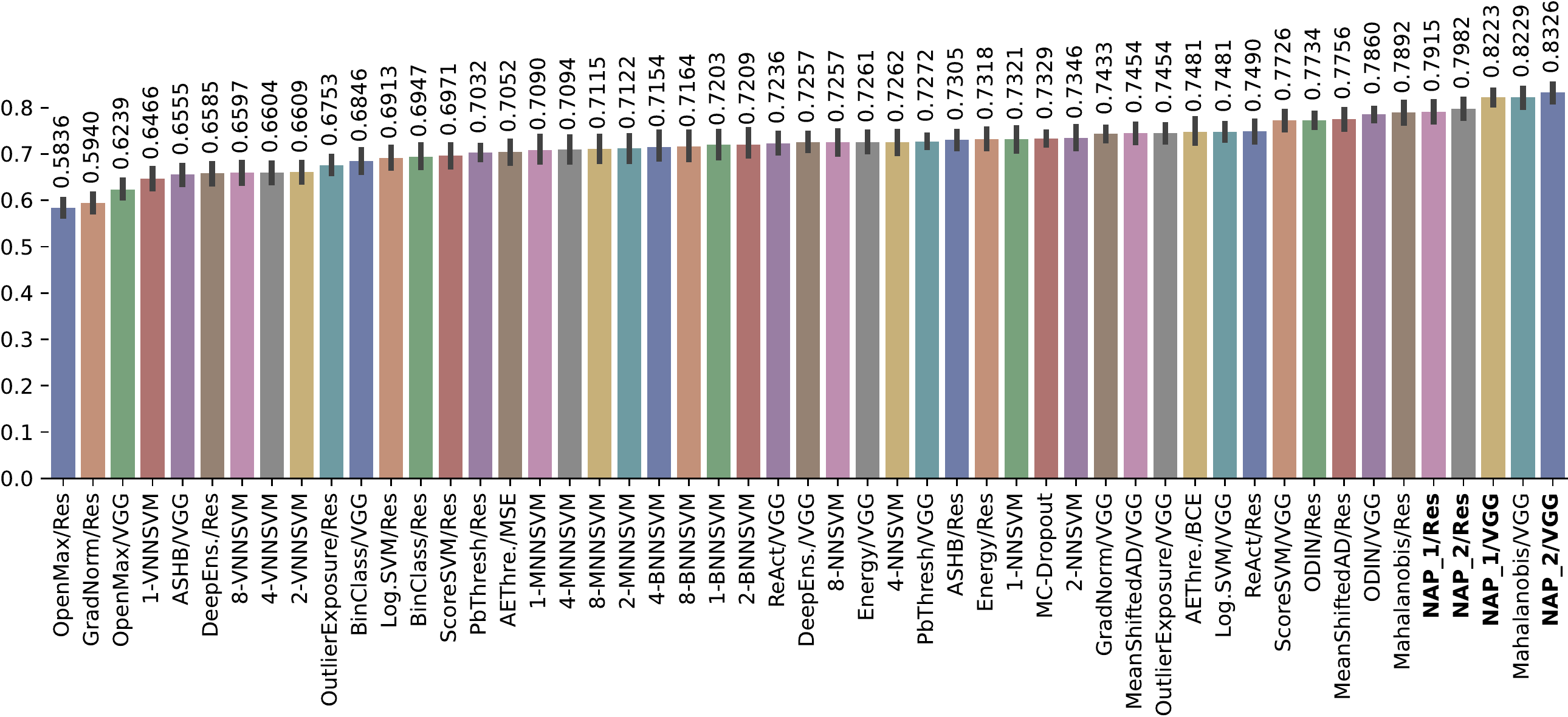}
    \caption{Accuracy}
    \label{fig:od-test acc}
    \end{subfigure}
    \caption{Average evaluation results of the proposed NAP approach with various configurations with state-of-the-art OOD detection methods over all datasets: (a) AUROC and (b) accuracy. The suffix /VGG or /Res indicates the network architecture. }
\end{figure*}

 \begin{table}[]
 \caption{Comparison of the mean processing time per single image for state-of-the-art OOD detectors. The results of our method are highlighted in bold.}
\label{table:time efficiency}
\scriptsize
\begin{center}
\begin{tabular}{llc} \hline \hline
\ Method & \multicolumn{2}{c}{Time [s]}  \\
\hline
\hline
 & VGG & ResNet \\
 \cline{2-3}
 ASH-B & 0.004 & 0.009 \\
 BinClass & 0.002  & 0.007 \\
 DeepEns. & 0.012 & 0.038 \\
 Energy & 0.003   & 0.008 \\
 GradNorm & 0.010  & 0.027 \\
 Log.SVM & 0.003  & 0.006 \\
 MC-Dropout & 0.003   & 0.007 \\
 Mahalanobis & 0.217 & 0.086 \\
 MeanShiftedAD & 0.231 & 0.314 \\
 
 ODIN & 0.010 & 0.028 \\
 OpenMax& 0.035 & 0.040  \\
 OutlierExposure & 0.003  & 0.008 \\
 PbThresh & 0.002  & 0.007  \\
 ReAct& 0.003 & 0.008 \\
 ScoreSVM & 0.002 & 0.006\\ 
 
\bf{Ours (NAP)} & \bf{  0.008} &  \bf{ 0.015} \\
 \hline
\end{tabular}
\end{center}
\end{table}
\noindent \textbf{Comparison with state-of-the-art OOD detection} Finally, we compared the best configurations established in the previous experiment with competitive methods proposed in the literature for OOD detection.  The obtained results are presented in Fig. \ref{fig:odtest auroc} using the AUROC metric and in Fig.  \ref{fig:od-test acc} for the accuracy metric.   
Each method in the presented comparison was tested over the same set of 308 experiments and then averaged, where all datasets were respectively used as train $D_{s}$, validation $D_{v}$, and test $D_{t}$.

The performed evaluation shows that for both considered metrics, the proposed binary NAP outperforms state-of-the-art OOD detection methods by a considerable margin. With an appropriate choice of hyperparameters, the NAP-based method can achieve a 3.4\% higher AUROC score than the second-best Mahalanobis on VGG architecture; 0.1 \% higher than the second-best ODIN on ResNET; 1.2\% higher accuracy on VGG than the second-best Mahalanobis;  1.1\% higher accuracy on ResNET than the second-best Mahalanobis.
This method is tightly dependent not only on the architecture of a classifier but also on how and how well the underlying classifier is trained. The hyperparameters are universally set for all architectures and $D_{s}$ training datasets for the purpose of fair evaluation. But they could be further optimized for each $D_{s}$ vs $D_{v}$ task. We suggest one of the following strategies as by default: (1) No of votes - 5; auto-configuration criterion - threshold. The strategy achieves the best AUROC value for both the architectures and the best and second-best accuracy. (2) No of votes - 9; auto-configuration criterion - validation accuracy. This yields the second-best AUROC value on ResNET (slightly lower than ODIN), but prevails in every other case, performing especially well on VGG. 
\par Additionally, we evaluated the performance of all methods and aggregated the results per dataset. These results are presented in Tab. \ref{tab: results_per_class}. While the proposed  NAP variants outperform state-of-the-art methods when the results are averaged for all datasets, we can observe that NAP may achieve slightly worse results for specific datasets.

\begin{table*}[!htb]
\scriptsize
\begin{center}
\caption{OD-test AUROC for every training dataset. The best results for all datasets and for each dataset are highlighted in bold. Note that the NAP\_2 version of our method is the majority voting system and the AUROC metric cannot be computed for it. \label{tab: results_per_class}
}
\begin{tabular}{llccccccc|ccccccc} 
\hline\noalign{\smallskip}
Model & Method & All & MNIST & FMNIST & CIFAR10 & CIFAR100 & STL10 & TImagenet & All & MNIST & FMNIST & CIFAR10 & CIFAR100 & STL10 & TImagenet \\
\noalign{\smallskip}
\hline
\noalign{\smallskip}
& & \multicolumn{7}{c}{ AUROC} & \multicolumn{7}{c}{ Accuracy}  \\ \hline \\
\multirow{15}{*}{VGG} & ASH-B@90 & 0.6935 & 0.9801 & 0.9029 & 0.5341 & 0.4971 & 0.7122 & 0.4991  & 0.6555 & 0.9297 & 0.7807 & 0.5343 & 0.5048 & 0.6363 & 0.5116\\
& BinClass & 0.7721 & 0.9802 & 0.8485 & 0.7251 & 0.5944 & 0.7737 & 0.6989 & 0.6846 & 0.9281 & 0.6826 & 0.6604 & 0.5664 & 0.6631 & 0.5955 \\
& DeepEns. & 0.7744 & 0.9878 & 0.8967 & 0.8410 & 0.6005 & 0.7774 & 0.5602 & 0.7257 & 0.9491 & 0.7992 & 0.7853 & 0.5594 & 0.6675 & 0.5935 \\
& Energy & 0.8148 & 0.9882 & 0.9512 & 0.8634 & 0.7448 & 0.8664 & 0.4996 & 0.7261 & 0.9470 & 0.8348 & 0.70.16 & 0.6546 & 0.70.80 & 0.4996 \\
& GradNorm & 0.7911 & 0.9801 & 0.8703 & 0.8398 & 0.6988 & 0.7086 & 0.6404 & 0.7433 & 0.9336 & 0.7963 & 0.7826 & 0.6459 & 0.6719 & 0.6211 \\
& Log.SVM & 0.8368 & 0.9793 & 0.9177 & 0.8227 & 0.7289 & 0.8693 & 0.7070 & 0.7481 & 0.9087 & 0.8003 & 0.7378 & 0.6215 & 0.7622 & 0.6588 \\
& MC-Dropout & 0.7979 & 0.9762 & 0.8635 & 0.8236 & 0.7205 & 0.7022 & 0.6837 & 0.7329 & 0.9285 & 0.7794 & 0.7520 & 0.6483 & 0.6228 & 0.6433  \\
& Mahalanobis & 0.8836 & \textbf{0.9999} & 0.9863 & 0.8872 & 0.6816 & 0.9104 & 0.8434 & 0.8229 & 0.9390 & 0.9213 & \textbf{0.8296} & 0.6747 & \textbf{0.8343} & 0.7425 \\
& MeanShiftedAD & 0.7995 & 0.9950 & 0.9666 & 0.8381 & 0.7700 & 0.7213 & 0.4961  & 0.7454 & 0.9745 & 0.8926 & 0.7473 & 0.6637 & 0.6832 & 0.4955 \\
& \bf{Ours (NAP\_1)} & \textbf{0.9137} & 0.9996 & \textbf{0.9836} & 0.8359 & 0.7720 & \textbf{0.9520} & \textbf{0.9289}  & 0.8223 & 0.9814 & 0.9125 & 0.7440 & 0.6768 & 0.8232 & 0.7761 \\
& \bf{Ours (NAP\_2)} & N/A & N/A & N/A & N/A & N/A & N/A & N/A & \textbf{0.8326} & \textbf{0.9901} & \textbf{0.9396} & 0.7401 & 0.6808 & 0.7954 & \textbf{0.8168} \\
& ODIN & 0.8650 & 0.9808 & 0.8883 & \textbf{0.9152} & \textbf{0.7841} & 0.8913 & 0.7494 & 0.7860 & 0.9335 & 0.8047 & 0.8209 & \textbf{0.6861} & 0.7844 & 0.6945 \\
& OpenMax& 0.6400 & 0.7249 & 0.5000 & 0.6395 & 0.6095 & 0.7638 & 0.6330 & 0.6239 & 0.6927 & 0.5000 & 0.5768 & 0.5857 & 0.7467 & 0.6604 \\
& OutlierExposure & 0.8420 & 0.9361 & 0.9075 & 0.8466 & 0.7520 & 0.8360 & 0.7730 & 0.7454 & 0.9166 & 0.8020 & 0.7375 & 0.6471 & 0.7229 & 0.6383 \\
& PbThresh & 0.7615 & 0.9371 & 0.7412 & 0.8245 & 0.7018 & 0.6936 & 0.6696 & 0.7272 & 0.9144 & 0.7311 & 0.7709 & 0.6426 & 0.6735 & 0.6281\\
& ReAct& 0.8120 & 0.9699 & 0.9556 & 0.8576 & 0.7423 & 0.8743 & 0.4988 & 0.7236 & 0.9299 & 0.8437 & 0.7008 & 0.6532 & 0.7043 & 0.4988 \\
& ScoreSVM & 0.8335 & 0.9873 & 0.9697 & 0.8357 & 0.6608 & 0.8781 & 0.6810 & 0.7726 & 0.9408 & 0.8989 & 0.7348 & 0.5923 & 0.7702 & 0.6880 \\

\midrule

\multirow{14}{*}{ResNet} & ASH-B@90 & 0.8021 & 0.9499 & 0.8482 & 0.8638 & 0.7048 & 0.8455 & 0.6262 & 0.7305  & 0.9037 & 0.7727 & 0.7333 & 0.6297 & 0.7183 & 0.6226\\
& BinClass & 0.7451 & 0.9947 & 0.7613 & 0.6791 & 0.6376 & 0.7001 & 0.6698 & 0.6947 & 0.9027 & 0.6703 & 0.6523 & 0.5974 & 0.6607 & 0.6657\\
& DeepEns. & 0.7214 & 0.9544 & 0.8512 & 0.7791 & 0.5423 & 0.6562 & 0.5432 & 0.6585 & 0.8475 & 0.7604 & 0.6878 & 0.5328 & 0.5900 & 0.5225 \\
& Energy & 0.8248 & 0.9762 & 0.9070 & 0.8940 & 0.6555 & 0.8238 & 0.7090 & 0.7318 & 0.9165 & 0.8095 & 0.7466 & 0.5722 & 0.7139 & 0.6311 \\
& GradNorm & 0.6311 & 0.7946 & 0.4771 & 0.4379 & 0.7415 & 0.3733 & 0.8494 & 0.5940 & 0.6744 & 0.5021 & 0.4861 & 0.6719 & 0.4252 & 0.7346\\
& Log.SVM & 0.7710 & 0.9292 & 0.8404 & 0.7315 & 0.7176 & 0.7230 & 0.6621  & 0.6913 & 0.8531 & 0.7330 & 0.6357 & 0.5811 & 0.6352 & 0.6813\\
& Mahalanobis & 0.8459 & 0.9959 & 0.9805 & 0.8485 & 0.6537 & 0.7579 & 0.8175 & 0.7892 & 0.9383 & 0.9148 & 0.7724 & 0.6465 & 0.7491 & 0.6997 \\
& MeanShiftedAD & 0.8427 & 0.9984 & 0.9733 & \textbf{0.9343} & \textbf{0.8161} & 0.6495 & 0.6588  & 0.7756 & 0.9739 & 0.9001 & \textbf{0.8447} & \textbf{0.7031} & 0.6268 & 0.5836\\
& \bf{Ours (NAP\_1)} & 0.8527 & \textbf{0.9998} & \textbf{0.9875} & 0.8482 & 0.6523 & 0.7847 & 0.8253 & 0.7915 & 0.9892 & 0.9294 & 0.7605 & 0.5798 & 0.7368 & 0.7314 \\
& \bf{Ours (NAP\_2)} & \textbf{0.8567} & 0.9996 & 0.9874 & 0.8355 & 0.6614 & 0.8053 & 0.8328 & \textbf{0.7982}& \textbf{0.9875} & \textbf{0.9422} & 0.7557 & 0.5969 & 0.7397 & 0.7419\\
& ODIN & 0.8561 & 0.9247 & 0.8337 & 0.9231 & 0.7597 & 0.8886 & 0.8316 & 0.7734 & 0.8630 & 0.7519 & 0.8387 & 0.6720 & 0.7868 & \textbf{0.7474}  \\
& OpenMax & 0.6054 & 0.6755 & 0.5000 & 0.6388 & 0.5613 & 0.6271 & 0.6430 & 0.5836 & 0.6531 & 0.5000 & 0.6064 & 0.5413 & 0.5847 & 0.6218\\
& OutlierExposure & 0.7774 & 0.9172 & 0.8436 & 0.8228 & 0.6058 & 0.7510& 0.7286 & 0.6753 & 0.8593 & 0.7067 & 0.6876 & 0.5510 & 0.6155 & 0.6197\\
& PbThresh & 0.7615 & 0.9170 & 0.8341 & 0.8453 & 0.5833 & 0.7540 & 0.6537 & 0.7032 & 0.8704 & 0.7652 & 0.7814 & 0.5264 & 0.6850 & 0.6054 \\
& ReAct & 0.8462 & 0.9825 & 0.9230 & 0.8961 & 0.6821 & \textbf{0.9064} & 0.7141 & 0.7490 & 0.9296 & 0.8091 & 0.7516 & 0.5727 & \textbf{0.8164} & 0.6318\\
& ScoreSVM & 0.8085 & 0.9902 & 0.8041 & 0.7679 & 0.6883 & 0.6528 & \textbf{0.8982} & 0.6971 & 0.8962 & 0.7194 & 0.6367 & 0.60.01 & 0.5987 & 0.6916 \\

\midrule

\multirow{10}{*}{---} & 1-NNSVM & \textbf{0.7956} & 0.9975 & 0.9864 & \textbf{0.8226} & \textbf{0.7813} & 0.6683 & 0.4921 & 0.7321 & 0.9735 & 0.8879 & 0.6899 & \textbf{0.6490} & 0.6295 & 0.5265\\
& 1-BNNSVM & 0.6680 & 0.9996 & \textbf{0.9984} & 0.4787 & 0.4920 & 0.4833 & 0.4623  & 0.7203 & 0.9919 & \textbf{0.9763} & 0.5686 & 0.5596 & 0.6174 & 0.5443 \\
& 1-MNNSVM & 0.6678 & 0.9995 & 0.9962 & 0.4743 & 0.4945 & 0.4824 & 0.4648 & 0.7090 & 0.9853 & 0.9577 & 0.5536 & 0.5519 & 0.5947 & 0.5430 \\
& 1-VNNSVM & 0.6165 & 0.9270 & 0.7904 & 0.4919 & 0.4916 & 0.4808 & 0.4519 & 0.6466 & 0.8679 & 0.7156 & 0.5751 & 0.5501 & 0.5947 & 0.5448\\
& 8-NNSVM & 0.7865 & 0.9963 & 0.9845 & 0.8178 & 0.7499 & 0.6661 & 0.4818 & 0.7257 & 0.9721 & 0.8802 & 0.6948 & 0.6363 & 0.6049 & 0.5277 \\
& 8-BNNSVM & 0.6629 & 0.9997 & 0.9983 & 0.4702 & 0.4727 & 0.4821 & 0.4618 & 0.7164 & 0.9901 & 0.9723 & 0.5755 & 0.5344 & 0.6193 & 0.5472 \\
& 8-MNNSVM & 0.6641 & 0.9995 & 0.9965 & 0.4661 & 0.4839 & 0.4814 & 0.4609 & 0.7115 & 0.9831 & 0.9569 & 0.5601 & 0.5498 & 0.60.33 & 0.5505 \\
& 8-VNNSVM & 0.6155 & 0.9424 & 0.7860 & 0.4862 & 0.4820 & 0.4806 & 0.4495 & 0.6597 & 0.8790 & 0.70.64 & 0.6111 & 0.5682 & 0.6172 & 0.5534 \\
& AEThre./BCE & 0.7315 & \textbf{1.0000} & 0.8846 & 0.6802 & 0.5434 & 0.6599 & \textbf{0.5898} & \textbf{0.7481} & \textbf{0.9982} & 0.8408 & \textbf{0.7423} & 0.5930 & \textbf{0.6598} & \textbf{0.6308}\\
& AEThre./MSE & 0.7739 & 0.9999 & 0.9650 & 0.7828 & 0.7501 & \textbf{0.7251} & 0.4100 & 0.70.52 & 0.9296 & 0.8230 & 0.6538 & 0.6346 & 0.6222 & 0.5340 \\

\hline
\end{tabular}
\end{center}

\label{table:Od-test AUROC per training dataset}
\end{table*}

\section{Conclusions}

In this work, we introduced a novel method for the detection OOD data for DNNs designed for image recognition problems. Extensive experiments allow the optimal configuration to be found for NAP-based convolutional neural networks that can efficiently identify the OOD samples. The proposed approach significantly outperforms the results achieved by state-of-the-art methods. Moreover, the proposed approach works on binary vectors and is characterized by very low computational complexity due to the binary representation of the extracted features. We believe that this work will inspire further examination of the theoretical properties of NAPs in the context of OOD detection.

\section*{Acknowledgements}
\noindent This work was supported by the National Centre for Research and Development under the project LIDER/51/0221/L-11/19/NCBR/2020. 

\clearpage
\clearpage

\def\IEEEbibitemsep{0pt plus .5pt}

\bibliographystyle{ieee_fullname}
\bibliography{egbib}

\end{document}